# Sentiment Classification of Gaza War Headlines: A Comparative Analysis of Large Language Models and Arabic Fine-Tuned BERT Models

**Amr Eleraqi**

MSc Candidate, Faculty of Graduate Studies and Research

Cairo University, Egypt

**Hager H. Mustafa**

Independent Researcher, Anmat Media

**Prof. Abdul Hadi N. Ahmed**

Professor, Faculty of Graduate Studies and Research

Cairo University, Egypt

## Abstract

This study examines how different artificial intelligence architectures interpret sentiment in conflict-related media discourse, using the 2023 Gaza War as a case study. Drawing on a corpus of 10,990 Arabic news headlines (Eleraqi 2026), the research conducts a comparative analysis between three large language models and six fine-tuned Arabic BERT models. Rather than evaluating accuracy against a single human-annotated gold standard, the study adopts an epistemological approach that treats sentiment classification as an interpretive act produced by model architectures. To quantify systematic differences across models, the analysis employs information-theoretic and distributional metrics, including Shannon Entropy, Jensen–Shannon Distance, and a Variance Score measuring deviation from aggregate model behavior. The results reveal pronounced and non-random divergence in sentiment distributions. Fine-tuned BERT models—particularly MARBERT—exhibit a strong bias toward neutral classifications, while LLMs consistently amplify negative sentiment, with LLaMA-3.1-8B showing near-total collapse into negativity. Frame-conditioned analysis further demonstrates that GPT-4.1 adjusts sentiment judgments in line with narrative frames (e.g., humanitarian, legal, security), whereas other LLMs display limited contextual modulation. These findings suggest that the choice of model constitutes a choice of interpretive lens, shaping how conflict narratives are algorithmically framed and emotionally evaluated. The study contributes to media studies and computational social science by foregrounding algorithmic discrepancy as an object of analysis and by highlighting the risks of treating automated sentiment outputs as neutral or interchangeable measures of media tone in contexts of war and crisis.

**Keywords:** War journalism; Media framing; Automated sentiment analysis; Arabic media; Gaza War; Algorithmic bias; Computational discourse analysis

# المستخلص

تبحث هذه الدراسة في كيفية تفسير نماذج مختلفة من الذكاء الاصطناعي للمشاعر في الخطاب الإعلامي المرتبط بالنزاعات المسلحة، وذلك من خلال دراسة حالة تغطية حرب غزة عام 2023. وتعتمد الدراسة على تحليل مجموعة بيانات تضم 10,990 عنوانًا خبريًا باللغة العربية، وتُجري مقارنة منهجية بين ثلاثة نماذج لغوية كبيرة (LLMs) وستة نماذج عربية من عائلة BERT جرى تدريبها وضبطها خصيصًا لتحليل المشاعر. وعلى خلاف الدراسات التي تنطلق من افتراض وجود معيار بشري واحد يمثّل «التصنيف الصحيح»، تنطلق هذه الدراسة من منظور إبستمولوجي يعتبر تصنيف المشاعر عملية تفسيرية تتأثر ببنية النموذج، وأهداف تدريبه، ونوع البيانات التي تعلّم منها. وبناءً على ذلك، لا تُقاس الفروق بين النماذج بوصفها أخطاء، بل بوصفها أنماطًا دلالية مختلفة في قراءة الخطاب الإعلامي. ولرصد هذه الفروق بصورة منهجية، توظّف الدراسة مجموعة من المقاييس التوزيعية والمعلوماتية، من بينها إنتروبيا شانون، ومسافة جنسن–شانون، ومؤشر التباين، وذلك بهدف قياس درجة الاختلاف والانحراف بين سلوكيات النماذج المختلفة.

تكشف النتائج عن وجود تباينات واضحة وغير عشوائية في كيفية تصنيف المشاعر بين النماذج محل الدراسة. إذ تُظهر نماذج BERT العربية المضبوطة دقيقًا—وخاصة نموذج—MARBERT ميلًا قويًا نحو تصنيف غالبية العناوين الإخبارية بوصفها محايدة، بما يعكس حذرًا تقييميًا مرتفعًا. في المقابل، تميل النماذج اللغوية الكبيرة إلى تصعيد التقييم السلبي بصورة منهجية، ويظهر هذا الاتجاه بأقصى درجاته في نموذجLLaMA-3.1-8B ، الذي يصنّف معظم العناوين تقريبًا ضمن الفئة السلبية. كما يُبرز التحليل المشروط بالإطار السردي أن نموذج GPT-4.1 يمتلك قدرة أعلى على مواءمة الحكم الشعوري مع طبيعة الإطار الخطابي المستخدم في العنوان الإخباري، سواء كان إنسانيًا أو قانونيًا أو أمنيًا أو سياسيًا، في حين تُظهر نماذج لغوية كبيرة أخرى قدرة محدودة على هذا النوع من التكيّف السياقي.

وتخلص الدراسة إلى أن اختيار نموذج تحليل المشاعر لا يُعد قرارًا تقنيًا محايدًا، بل يمثل اختيارًا لعدسة تفسيرية تؤثر بصورة مباشرة في كيفية قراءة الخطاب الإعلامي للحروب والنزاعات. وبذلك تسهم هذه الدراسة في حقل دراسات الإعلام والعلوم الاجتماعية الحاسوبية من خلال إعادة توجيه الانتباه من «دقة التصنيف» إلى «اختلاف التفسير»، والتنبيه إلى المخاطر المعرفية المترتبة على التعامل مع نواتج تحليل المشاعر الآلي بوصفها مؤشرات موضوعية أو قابلة للتبادل في سياقات النزاع والحروب.

## Introduction

Media coverage of wars and political crises has long been a central concern in communication and media research. Scholars have consistently demonstrated that news media do not merely report events but actively construct narratives that shape how conflicts are understood, evaluated, and morally interpreted (McCombs and Shaw 1972). Through established processes such as agenda-setting and framing, media outlets exert significant influence over which aspects of a conflict are emphasized, which actors are foregrounded, and how concepts like responsibility, suffering, and legitimacy are communicated to audiences (Entman 1993).

Within this tradition, the evaluative tone—often operationalized as sentiment—plays a crucial role in the analysis of conflict reporting. The distinction between negative, neutral, or positive coverage is not a trivial one; it reflects the framing of violence, humanitarian suffering, political decisions, and legal accountability. However, in the context of war reporting, sentiment is rarely expressed through overt emotional language alone. Headlines, in particular, may remain lexically factual while implicitly conveying evaluation through context, framing, and assumed background knowledge.

In recent years, the adoption of automated sentiment analysis has provided a powerful tool for studying media discourse at scale, offering the promise of efficiency and consistency. This has enabled researchers to analyze thousands of news items across various outlets and time periods. Consequently, sentiment scores are increasingly treated as objective, interchangeable indicators of media tone in studies of political communication, conflict, and public opinion.

Yet, this growing reliance on algorithmic sentiment classification raises a fundamental, yet underexplored, question in media studies: Do different computational models interpret the emotional tone of the same news coverage in a consistent manner? More specifically, when applied to the complex and contested domain of conflict reporting, do automated systems converge on a shared understanding of evaluative meaning, or do they reproduce distinct and potentially competing interpretations of the same discourse?

This question is particularly pressing in the context of contemporary wars, where media narratives are highly contested and politically consequential. In such environments, the very notion of neutrality can be interpreted in multiple ways: as professional detachment, as moral distancing, or as implicit alignment. Furthermore, negative evaluations often stem not from emotional language but from the recognition of violence, harm, or injustice embedded within otherwise factual reporting.

Most existing research in sentiment analysis has focused on improving classification accuracy by comparing model outputs to human-annotated datasets. While this approach is valuable, it rests on the assumption that sentiment represents a stable "ground truth" that models should approximate. From a media studies perspective, however, evaluative meaning is inherently interpretive. Different readers—and different analytical frameworks—may reasonably disagree on how a headline positions an event or actor.

This study adopts an alternative, critical perspective. Rather than treating disagreement between models as a measure of error, we treat such divergence as analytically meaningful. The paper systematically compares sentiment classifications produced by two broad families of AI systems: specialized Arabic BERT models fine-tuned for sentiment analysis, and large language models (LLMs) designed for general-purpose language understanding. By examining how these systems diverge in their assessments of the same Arabic news headlines, the study aims to reveal how different computational "reading strategies" reorganize the emotional structure of conflict discourse.

The empirical focus is a corpus of 10,990 Arabic news headlines related to the Gaza War (Eleraqi 2026). This case provides a particularly suitable context for analysis, as coverage of the conflict combines humanitarian, legal, security, political, and historical frames within a highly polarized media environment. The study analyzes sentiment distributions across all models and, for LLMs, specifically examines how sentiment judgments vary across pre-defined narrative frames.

Methodologically, the study avoids imposing a single human-coded benchmark as the definitive standard. Instead, it compares model outputs using distributional and information-theoretic measures to identify systematic patterns of convergence and divergence. By doing so, it reframes automated sentiment analysis as a form of mediated interpretation rather than a neutral measurement tool. By bridging computational analysis with concepts from media and discourse studies, this paper contributes to ongoing debates about the role of algorithms in shaping knowledge about war and crisis. It demonstrates that the choice of a sentiment model is not merely a technical decision but a substantive one, with direct implications for how media discourse is analyzed, summarized, and ultimately understood.

## Related Work

In media and communication research, large-scale monitoring of news coverage has long been anchored in content analysis—a systematic way to describe and interpret media texts using a defined coding scheme, so that patterns can be compared across outlets, topics, and time (Krippendorff 2019).

Core monitoring questions (e.g., which issues receive the most attention, which actors are foregrounded, and what kinds of evaluations are attached to them) map directly onto central theories such as agenda-setting and framing. Agenda-setting scholarship argues that the prominence of issues in news coverage helps structure perceived issue importance, linking the distribution of media attention to public issue salience (McCombs and Shaw 1972).

Framing research further explains how meaning is organized in news discourse: by selecting certain aspects of reality and making them more salient in ways that promote particular problem definitions, causal interpretations, moral evaluations, and policy remedies (Entman 1993).

Computational approaches extend these traditions by enabling content-analytic measurement at scale, but they also shift what "measurement" entails. Work on text-as-data emphasizes that automated methods do not simply substitute for human coding; instead, they embed modeling assumptions that must be theorized and validated, because different methods can generate systematically different representations of the same political text (Grimmer and Stewart 2013).

This point is especially relevant for sentiment analysis, which is widely used in political communication as a scalable proxy for evaluative tone, yet can vary substantially depending on whether it is operationalized through dictionaries, supervised labels, or hybrid designs (Haselmayer and Jenny 2017).

In Arabic news, these methodological sensitivities are amplified by the language's orthographic and morphological properties, and by the common need for normalization steps to reduce sparsity and encoding variation prior to modeling (Habash 2010)[1].

Against this background, the present study treats sentiment classification not as a single "true score," but as an interpretive output produced by model architectures and their underlying training assumptions. By comparing the output distributions of encoder-based Arabic BERT models (optimized for bidirectional contextual representation) to LLM-based

1 Arabic normalization refers to preprocessing steps (e.g., orthographic unification, diacritic removal) used to reduce lexical sparsity in computational models.

classifications, we operationalize algorithmic discrepancy as an object of analysis—using disagreement patterns to show how different computational "lenses" can reorganize the evaluative structure of conflict news and, in turn, shape the analytical conclusions drawn from large-scale media data.

## 1. Sentiment Analysis in Arabic NLP

Early Arabic sentiment analysis typically followed two "classic" NLP pipelines: lexicon/rule-based scoring (counting polarity terms, handling negation, and applying handcrafted rules) and supervised corpus-based classifiers built on sparse features such as bag-of-words and n-grams (Mulki et al. 2017; Boudad et al. 2018).

Diagram 1. The Evolution of Arabic Sentiment Analysis with NLP

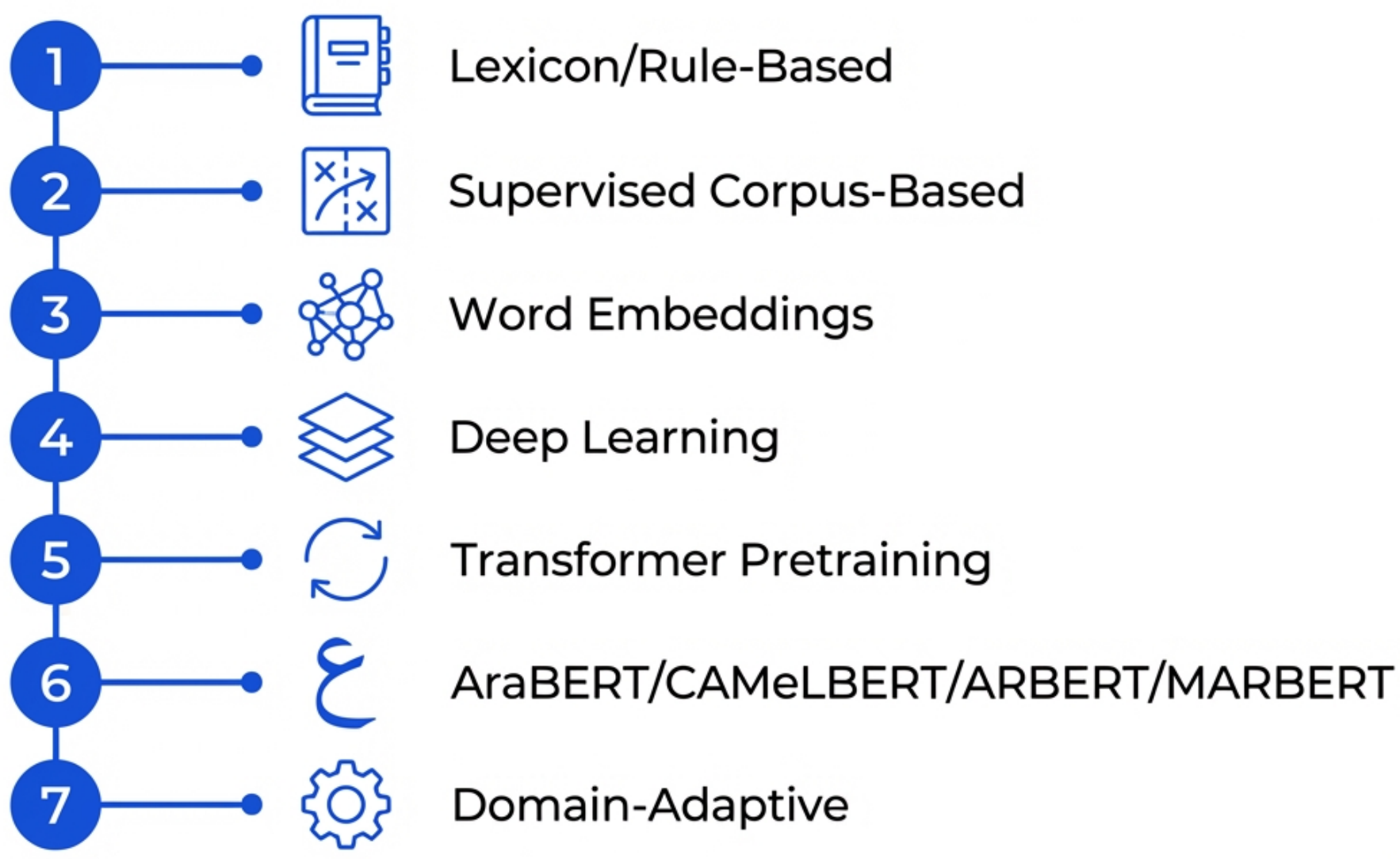

Source: Developed by the authors to trace the historical development of Arabic NLP methodologies.

In Arabic, both pipelines are especially brittle because tokenization and feature matching are sensitive to morphological richness, orthographic ambiguity, and dialectal variation, which increases sparsity and weakens cross-domain generalization (Habash 2010; Mulki et al. 2017).

The next methodological “school” reduced this sparsity by replacing discrete word counts with distributed representations (word embeddings), enabling neural models to learn sentiment cues from continuous vectors rather than manually engineered features (Mikolov et al. 2013).

This, in turn, accelerated the adoption of deep learning classifiers for sentence-level tasks (e.g., CNN-style text classifiers), which improved performance by learning task-relevant patterns directly from data (Kim 2014; Boudad, Faizi, and Thami 2018).

The major inflection point, however, came with Transformer-based pretraining (BERT), where models learn contextual representations that condition on both left and right context—an inductive bias that is particularly effective for short-text classification such as headlines (Devlin et al. 2019).

For Arabic specifically, Arabic-pretrained Transformers such as AraBERT, CAMeLBERT, and ARBERT/MARBERT were introduced to better capture Arabic’s linguistic properties by pretraining on large Arabic corpora and, in several cases, explicitly accounting for MSA vs. dialectal data regimes (Antoun, Baly, and Hajj 2020; Inoue et al. 2021; Abdul-Mageed, Elmadany, and Nagoudi 2021).

In parallel, research on continued/domain-adaptive pretraining has shown that adapting a pretrained Transformer to in-domain text can yield systematic gains, motivating domain-focused models for specialized narratives (e.g., conflict and political violence) (Gururangan et al. 2020; ; Ke et al. 2022).

Against this trajectory, our study uses Arabic BERT-family checkpoints as a linguistically specialized baseline for headline-level sentiment, then contrasts them with LLM behavior to examine how different training objectives and data exposures produce systematic differences in interpretive sentiment in conflict reporting.

## 2. Sentiment Analysis of Geopolitical and Conflict Discourse

The application of sentiment analysis to political science and media studies is a growing field, offering scalable methods to quantify evaluative tone and track aggregate public opinion signals (Ceron, Curini, and Iacus 2015).

Analyzing conflict discourse presents unique challenges, as the language is often highly charged, metaphorical, and subject to propaganda, making sentiment classification a non-trivial task that goes beyond simple positive/negative polarity (OpenAI 2023).

Studies analyzing media coverage of conflicts, such as the Syrian civil war and the recent Israel-Gaza conflict, have highlighted how sentiment-bearing terms are strategically used to frame actors and events, demonstrating that sentiment analysis can be a powerful tool for deconstructing media narratives (Abuasaker et al. 2025; Almutrash and Abudalfa 2024).

Our research contributes to this subfield by focusing on the highly contested information environment of the Gaza War, a context where sentiment is a key component of the narrative battle.

### 3. Comparative Analyses of LLMs and Fine-Tuned Models

The recent emergence of powerful LLMs has raised a critical question: do these massive, general-purpose models render smaller, task-specific fine-tuned models obsolete? A growing body of research is dedicated to answering this question. Studies comparing ChatGPT-4 with fine-tuned BERT baselines find that LLMs can be competitive for Arabic dialect sentiment analysis Hannani, Soudi, and Van Laerhoven 2024).

.However, this performance is not uniform across all languages or domains. Other studies have pointed out the limitations of LLMs, including their susceptibility to prompt-engineering, higher computational cost, and potential for generating factually incorrect explanations (Bommasani et al. 2021; Liu et al. 2021; Huang et al. 2023).

Our work directly engages with this ongoing debate by providing a head-to-head comparison in a non-English, high-stakes domain, focusing not just on classification output but also on the qualitative nature of the models' "reasoning" through an exploratory analysis of their disagreements.

## Research Design and Methods

### 1. Data Collection and Corpus Description

collected from a wide range of regional and international news outlets over the study period. The corpus includes headlines published by major Arabic and global media organizations, allowing for variation in editorial orientation, political alignment, and geographic origin. This diversity is essential for avoiding outlet-specific bias and for capturing the plural and contested nature of war-related media discourse.

Headlines were selected as the unit of analysis because they represent a highly condensed form of journalistic discourse. As the primary point of audience exposure, headlines play a critical role in framing events, signaling evaluative stance, and shaping first impressions of conflict narratives. Their compact structure also makes them a stringent test case for automated sentiment analysis, as evaluative meaning is often conveyed implicitly through framing choices rather than explicit emotional language.[2]

To construct the corpus, headlines were retrieved using five Arabic query keywords designed to capture major actors and recurring storylines in Gaza war coverage: "غزة" (Gaza), "أسرى" (captives/prisoners), "حماس" (Hamas), "الجيش الإسرائيلي" (the Israeli army), and "كتائب القسام" (al-Qassam Brigades). These keywords were selected to capture a broad spectrum of humanitarian, military, political, and organizational dimensions of the conflict. The resulting dataset is semi-balanced across keyword streams, with the following distribution: Gaza (2,550), captives (2,335), Hamas (2,231), Israeli army (2,029), and al-

[2] Headlines are treated here not merely as summaries of article content, but as *framing devices* that encode evaluative cues through lexical selection, actor positioning, and contextual implication—features central to media framing analysis.

Qassam Brigades (1,845). This relative balance reduces the risk that observed sentiment patterns are driven primarily by topical overrepresentation rather than by differences in model interpretation, thereby strengthening the validity of cross-model comparisons in subsequent analyses.

Figure 1. Distribution of Headlines Across the Five Key Terms

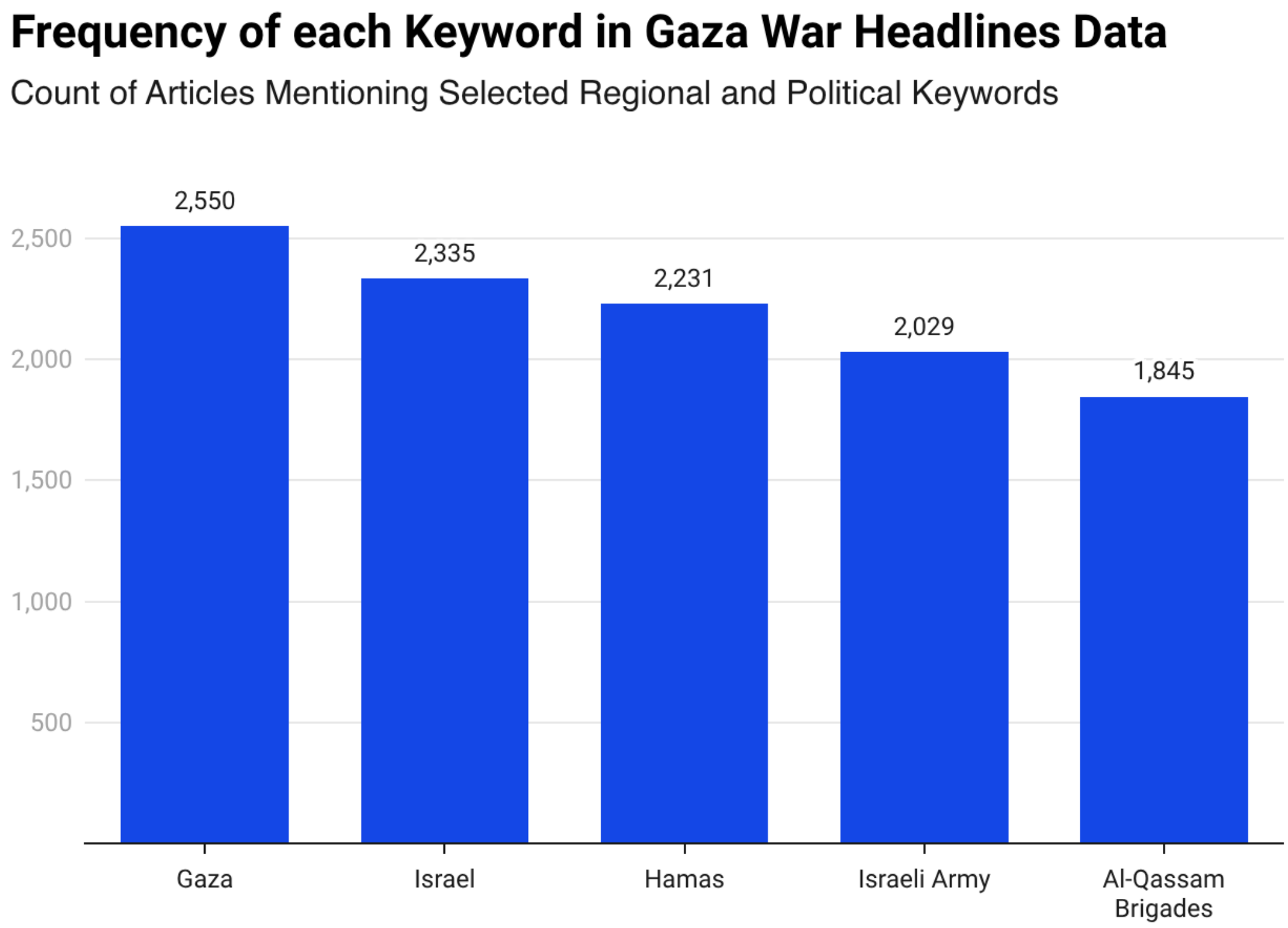

Source: Compiled by the authors based on the retrieved corpus data.

We further characterize the linguistic profile of the inputs by analyzing headline length using cleaned word count (i.e., the number of tokens after applying the preprocessing pipeline described in the next subsection). Overall, headline lengths are concentrated in the short-text regime: most headlines fall approximately within the 6–11 word range, with a right-skewed tail extending to about 37 words. This distribution indicates that the task primarily tests contextual interpretation in short, information-dense texts rather than long-sequence handling, a setting in which subtle framing cues must be inferred from limited lexical material. **As such, differences in model behavior are less likely to reflect capacity constraints and more likely to reflect divergent interpretive strategies applied to the** **same minimal textual evidence.** The following diagram illustrates the distribution of headline word counts.

Figure 2. Distribution of Headline Word Counts

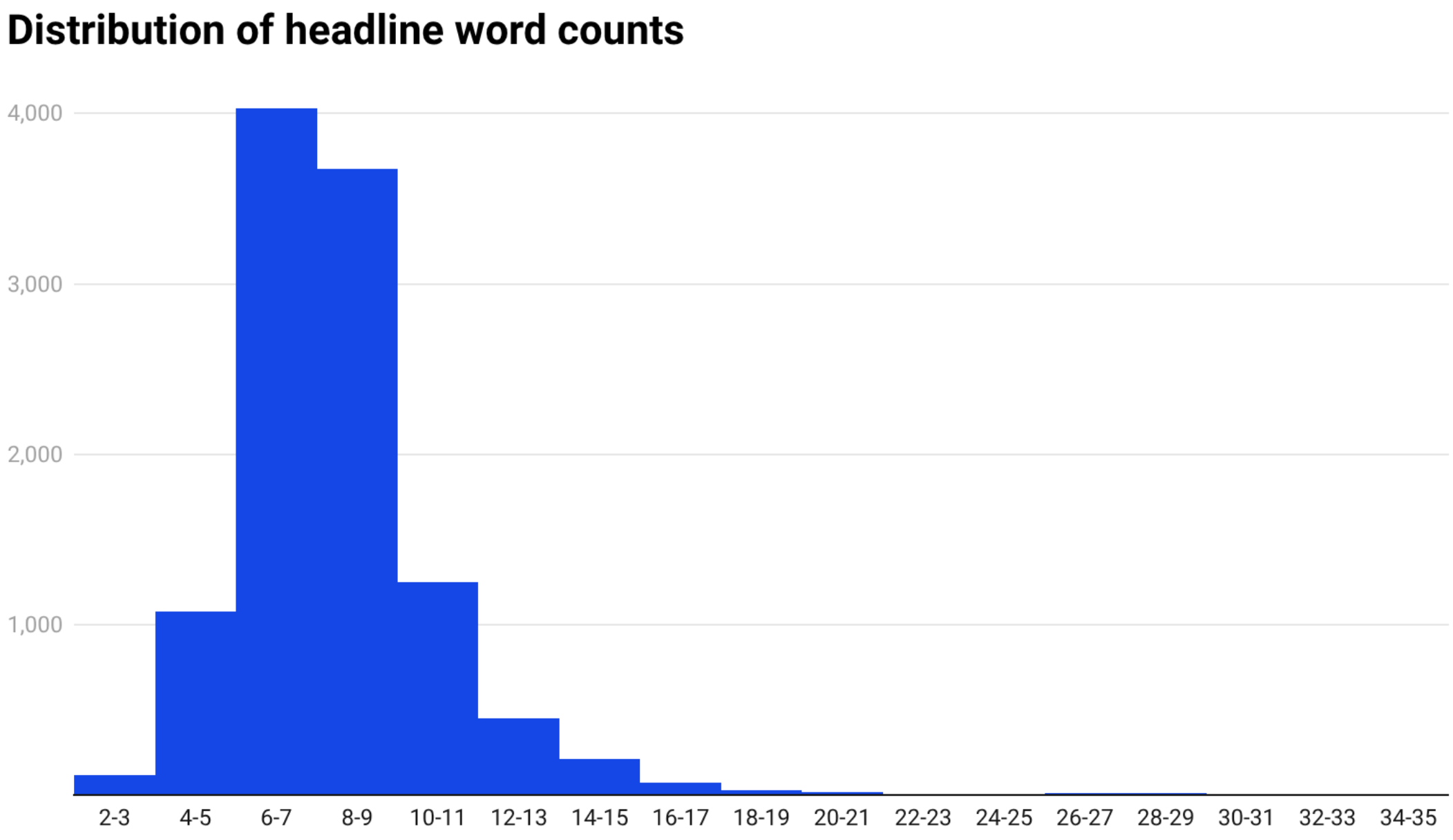

Source: Author's calculations based on the study dataset.

Finally, headline lengths are broadly comparable across keyword subsets. Mean cleaned length ranges narrowly from 8.6 to 9.5 words (Gaza: 9.5; al-Qassam Brigades: 9.2; Israeli army: 9.0; Hamas: 8.7; Captives: 8.6), and the following table summarizes show similar medians and dispersion across groups with occasional long outliers.

**Table 1. Descriptive Statistics of Headline Length by Query Keywords**

| Keyword (Arabic) | Keyword (English) | Mean Length (words) | Median | Std. Dev. | Min | Max |
|---|---|---|---|---|---|---|
| إسرائيل | Israel | 8.6 | 8.0 | 2.6 | 3 | 34 |
| الجيش الإسرائيلي | Israeli Army | 9.0 | 8.0 | 2.6 | 3 | 32 |
| حماس | Hamas | 8.7 | 9.0 | 2.3 | 3 | 35 |
| غزة | Gaza | 9.5 | 9.0 | 3.5 | 3 | 37 |
| كتائب القسام | Al-Qassam Brigades | 9.2 | 9.0 | 2.0 | 3 | 28 |
| Overall | Overall | 9.0 | 9.0 | 2.7 | 3 | 37 |

Source: Author’s calculations based on the study dataset.

**2. Text Preprocessing and Normalization**

All news headlines were subjected to a rigorous, rule-based preprocessing pipeline designed to standardize textual input while preserving sentiment-bearing lexical information. The primary objective of this preprocessing stage was to minimize linguistic noise and orthographic variability, thereby enabling the models to focus on the core semantic content of the headlines.

Headline cleaning was implemented as a deterministic pipeline and applied uniformly to the raw headline text prior to model inference. Each headline underwent the following sequential operations:

- First, trailing outlet or source identifiers were removed by splitting the headline at the final hyphen character and retaining only the left-hand segment. This step was intended to prevent media outlet names from influencing downstream textual representations.
- Second, Arabic diacritics were removed to eliminate pronunciation-level variation that does not contribute to sentiment interpretation in written headlines.
- Third, common Arabic orthographic variants were normalized to reduce sparsity caused by alternative spellings. Specifically, the characters أ، إ، and آ were mapped to ا; ى was mapped to ي; ؤ to و; ئ to ي; and ة to ه. This normalization step ensured greater lexical consistency across the corpus.

- Fourth, explicit media outlet names were removed using a corpus-derived list of source strings extracted from the dataset's metadata. This step further reduced source-specific lexical leakage into the textual content.
- Fifth, all Latin characters and numerical digits were removed, followed by the elimination of punctuation through replacement of non-word and non-space characters with whitespace.
- Sixth, Arabic stopwords were removed using an externally maintained stopword list, and single-character tokens were discarded to eliminate non-informative units.
- Finally, excess whitespace was normalized by collapsing multiple spaces into a single space and trimming leading and trailing whitespace.

After preprocessing, headlines containing fewer than three tokens were excluded to avoid degenerate or semantically impoverished inputs. This filtering step resulted in a final dataset of 10,990 cleaned news headlines. The preprocessing pipeline was applied identically across all model families to ensure that observed performance differences reflect model behavior rather than discrepancies in text preparation.

Diagram 2. Preprocessing Steps of the Database

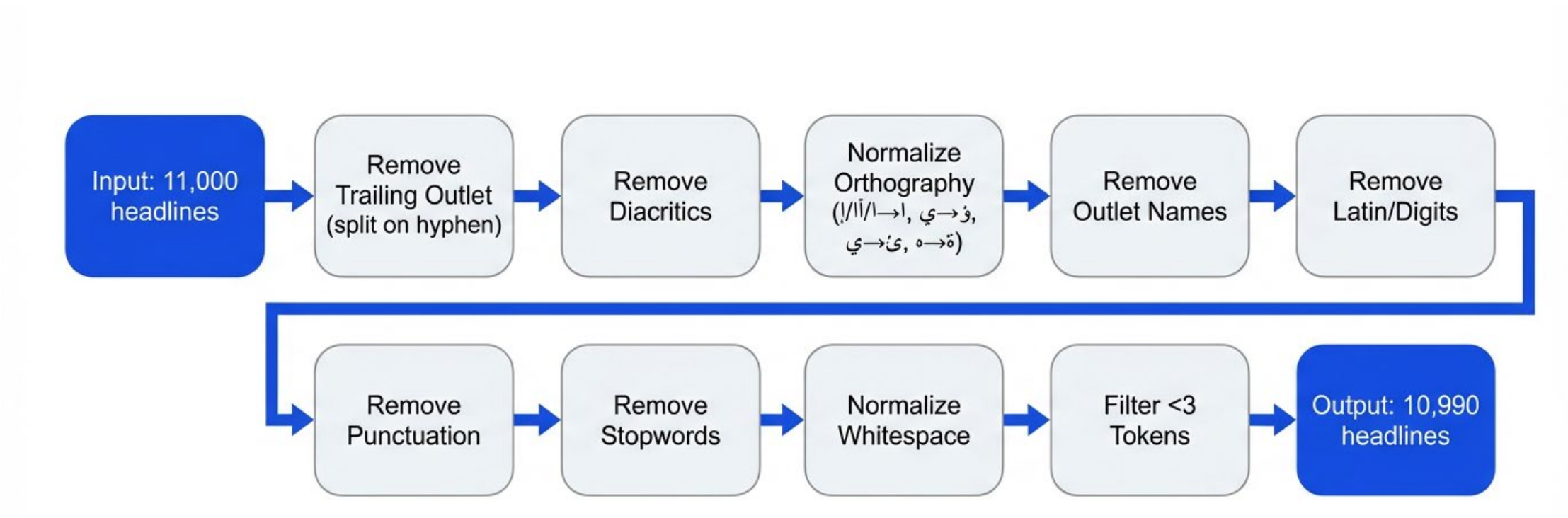

Source: Developed by the authors to illustrate the data cleaning workflow.

**Table 2. Representative Example Illustrating the Effect of the Preprocessing Pipeline**

| Row Index | Original Headline (Before) | Preprocessed Headline (After) | Preprocessing Modifications Applied to Sample Headline |
|---|---|---|---|
| 3615 | غزة: أثر مدمر للهجمات والحصار من جانب إسرائيل على الأشخاص ذوي الإعاقة Human Rights - Watch | غزه مدمر للهجمات والحصار جانب اسراييل علي الاشخاص ذوي الاعاقه | Removed punctuation and source attribution following the final dash. Normalized Arabic orthographic variants to a simplified canonical form. Retained named entities in normalized spelling without semantic alteration. Removed stopwords, including connectors and attribution-related phrases, to preserve core semantic content. |

Source: Generated from the preprocessing pipeline output.

Sentiment for this study is operationalized using a three-way polarity schema: positive, negative, and neutral. Crucially, the study relies solely on model-generated sentiment assignments for each headline. This design choice aligns with the paper's focus on algorithmic discrepancy, emphasizing how different architectures map the same input space into divergent sentiment spaces. The labels are treated as model outputs to be compared, rather than as a ground truth against which models are scored.

## 3. Computational Models for Sentiment and Framing

This study evaluates sentiment and framing performance across two distinct model families: general-purpose large language models (LLMs) and fine-tuned Arabic BERT-based classifiers. The inclusion of both families is intended to assess the extent to which large-scale contextual reasoning offered by contemporary LLMs provides measurable advantages over linguistically specialized models that have been explicitly trained on Arabic corpora.

The first group comprises six fine-tuned Arabic BERT models selected to represent a spectrum of Arabic NLP specialization, ranging from Modern Standard Arabic (MSA) and dialectal coverage to domain-specific continual pretraining. These models are encoder-based architectures optimized for classification tasks and have been widely adopted as strong baselines in Arabic sentiment and text analysis research. By contrasting these specialized models with generalist LLMs under a unified labeling scheme, the study aims to isolate the impact of architectural scale, pretraining domain, and linguistic specialization on both sentiment accuracy and framing sensitivity. This comparison allows us to assess whether fine-grained linguistic adaptation yields systematically different evaluative patterns than large-scale generalization.

The second group consists of three instruction-tuned LLMs—one proprietary and two open-weights—used as generalist comparators. These models are not specialized for Arabic sentiment analysis or media discourse tasks but instead rely on broad multilingual or cross-domain pretraining and instruction-following capabilities. Their inclusion enables an evaluation of how far general contextual understanding alone can substitute for task- or language-specific adaptation. Importantly, these models operate in a zero-shot regime, without exposure to domain-specific sentiment labels during fine-tuning.

Together, these nine models provide a comprehensive evaluation landscape that spans proprietary versus open-weights systems, generative versus encoder-based architectures, and general-purpose versus domain-adapted Arabic language models. This design makes it possible to disentangle technical performance from interpretive behavior, a distinction that is central to the study's media-analytical objectives.

**Diagram 3. A Summary of the Models Used in the Study**

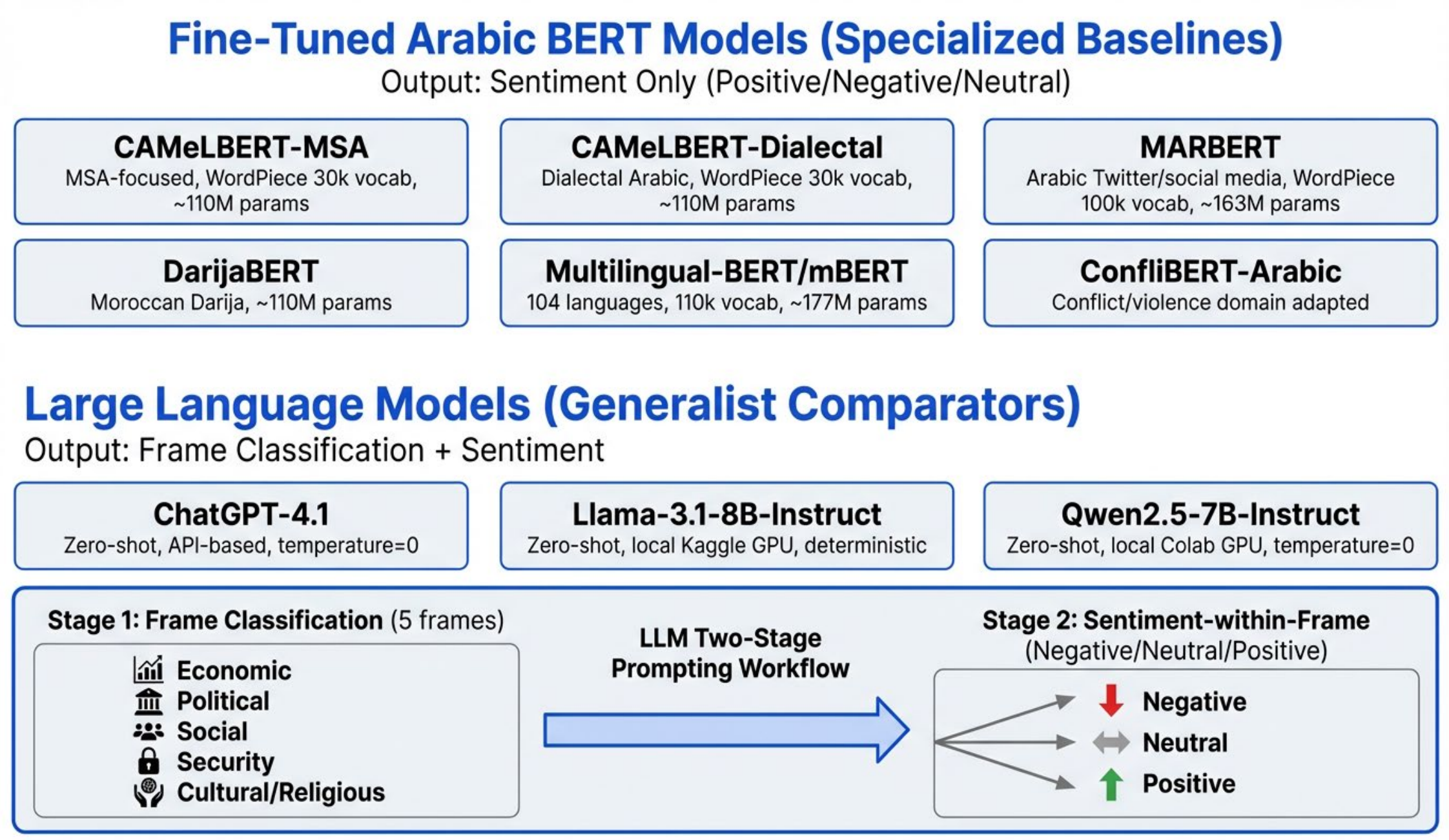

Source: Developed by the authors to categorize the experimental model families.

As the previous Diagram is showing. The study employs two distinct classes of models:

### 3.1. Fine-Tuned Arabic BERT Architectures

These models were selected to represent a spectrum of Arabic NLP specialization, from general-purpose to domain-specific.

**Table 3. Summary of Fine-Tuned Arabic BERT Models**

| Model | Pre-training data | Tokenizer / Vocab | Parameters |
|---|---|---|---|
| CAMeLBERT-MS | MSA-focused corpus | WordPiece, 30k vocab | ≈110M (BERT-Base sized) |
| CAMeLBERT-D | Dialectal Arabic corpus | WordPiece, 30k vocab | ≈110M (BERT-Base sized) |
| MARBERT | Arabic Twitter / social media | WordPiece, 100k vocab | ≈163M |
| DarijaBERT | Moroccan Darija sources (multiple) | 80K | ≈ 147M |
| Multilingual-BERT (mBERT) | Wikipedia in 104 languages | 110k vocab | ≈0.2B params |
| ConfliBERT-Ara | Continual pretraining for conflict/violence domain | — | — |

Source: CAMeLBERT model cards and training details (CAMeL Lab / Hugging Face); MARBERT documentation and ACL 2021 paper; DarijaBERT model statistics (Table 4); multilingual BERT (mBERT) model card; ConfliBERT-Arabic paper and Hugging Face organization documentation.

**Large Language Model (LLM) Frameworks**

To assess the robustness of sentiment and framing classification across different model families, we employed three large language models (LLMs) representing both proprietary and open-weights paradigms. All models were evaluated under a zero-shot setting and subjected to a unified two-stage inference protocol consisting of frame identification followed by sentiment classification conditioned on the predicted frame. This design ensured structural comparability across systems while allowing for differences in architectural capacity and training regimes.

***ChatGPT-4.1 (Generalist Proprietary Comparator)***

ChatGPT-4.1 was used as a state-of-the-art proprietary baseline representing contemporary general-purpose contextual reasoning. The model was accessed via the OpenAI API using the official Python SDK and the Responses endpoint. All production runs employed the model identifier gpt-4.1-2025-04-14 with temperature set to zero in order to minimize stochastic variation and ensure deterministic outputs.

Each headline was processed through a two-stage prompting pipeline. In the first stage, the model was instructed to identify exactly one dominant interpretive frame from a fixed set of five categories: Humanitarian, Security, Legal_Accountability, Political_Strategic, and Historical_Informational. In the second stage, the model assigned a sentiment label (Negative, Neutral, or Positive) explicitly conditioned on the previously predicted frame.

To enforce machine-readability and prevent post hoc interpretation, both stages required output in strict JSON format, validated against predefined schemas specifying allowed fields and values. The resulting outputs—frame label, frame explanation, sentiment label, and sentiment rationale—were stored per headline as {frame, frame_why, sentiment, sentiment_why} and used for both quantitative aggregation and qualitative frame analysis.

### *LLaMA-3.1-8B-Instruct (Open-Weights Comparator)*

LLaMA-3.1-8B-Instruct was employed as a strong open-weights comparator to evaluate non-proprietary model performance under the same analytical framework. The model was executed locally in a Kaggle GPU environment using the Hugging Face Transformers library (AutoModelForCausalLM). Inference was performed with deterministic decoding parameters (do_sample = False, num_beams = 1) to align with the zero-temperature setting used for ChatGPT-4.1.

The two-stage inference protocol mirrored that of the proprietary model at the level of decision structure: an initial frame classification followed by sentiment classification within the predicted frame. Primary inference requests required a one-line JSON object specifying the selected label and a short explanation or rationale.

Given the known tendency of decoder-only models to occasionally produce malformed or verbose outputs, we implemented a strict fallback mechanism. When JSON validation failed, the model was re-prompted to return only a single-letter code corresponding to the allowed frame (A–E) or sentiment (A–C). These codes were subsequently mapped to the canonical label set during post-processing. All fallback events were logged, and raw model completions were retained to ensure auditability.

To prevent extraction artifacts, only the newly generated completion tokens were decoded and analyzed, excluding the prompt text from the recorded output. This completion-only decoding strategy avoided inadvertent label leakage from the instruction template and ensured that stored labels reflected the model's generated decisions.

### *Qwen2.5-7B-Instruct (Open-Weights Comparator)*

Qwen2.5-7B-Instruct was included as an additional open-weights comparator and executed locally in a Colab GPU environment using the Hugging Face Transformers framework. The model was run in a zero-shot configuration with deterministic decoding (do_sample = False, temperature = 0.0) to maintain consistency with the other systems.

Inference followed the same two-stage structure used for ChatGPT-4.1 and LLaMA-3.1-8B-Instruct: first, selection of a single dominant frame from the fixed five-category set; second, sentiment classification conditioned on that frame. As with the other models, outputs were requested in constrained, machine-readable JSON format specifying both the label and a brief explanatory rationale.

When JSON validation failed at the frame-identification stage, a strict letter-only fallback (A–E) was enforced and mapped to the canonical frame labels during post-processing. All outputs were normalized into the unified three-way sentiment schema (Negative, Neutral, Positive) used across models to ensure direct comparability of output distributions.

## Inference Protocol and Label Harmonization

Across all three models, frame and sentiment labels were normalized into a shared taxonomy prior to analysis. Letter-coded fallbacks were mapped deterministically to their corresponding labels, and textual outputs were normalized case-insensitively. This harmonization ensured that observed differences in framing or sentiment distributions reflected model behavior rather than inconsistencies in label representation or output formatting.

All models were prompted using an identical decision structure and label taxonomy. Differences in outputs therefore reflect model behavior rather than prompt design or task formulation.

**Unified Prompt – Stage 1: Frame Classification**

You are a discourse analyst specializing in media framing.

Task:

Identify the PRIMARY interpretive frame used in the following Arabic news headline.

You MUST select exactly ONE frame from the list below:

- Humanitarian
- Security
- Legal_Accountability
- Political_Strategic
- Historical_Informational

Rules:

- Focus strictly on framing, not sentiment.
- Base your decision only on wording, emphasis, and implied responsibility.
- Do NOT add external facts, background knowledge, or personal opinions.

Output format:

Return ONLY valid JSON with exactly the following keys:

{

"frame": "<one of the five frames>",

"explanation": "<brief justification based on the headline wording>"

}

Do not include any text outside the JSON object.

**Unified Prompt – Stage 2: Sentiment Within Frame**

You are a media analyst.

Task:

Given the dominant frame identified for this headline, classify the sentiment expressed WITHIN THAT FRAME.

Sentiment labels:

- Negative: expresses harm, condemnation, loss, threat, or injustice within the frame.
- Neutral: descriptive or explanatory without evaluative judgment.
- Positive: expresses improvement, relief, success, or legitimacy within the frame.

Rules:

- The sentiment judgment MUST be conditioned on the provided frame.
- Do NOT reassess or change the frame.
- Do NOT add external facts or opinions.

Output format:

Return ONLY valid JSON with exactly the following keys:

{

"label": "<Negative | Neutral | Positive>",

"rationale": "<brief justification based on the headline wording>"

}

Do not include any text outside the JSON object.

**Unified Fallback Prompt – Frame (Letter-Only)**

Return ONLY one letter corresponding to the dominant frame:

A) Humanitarian

B) Security

C) Legal_Accountability

D) Political_Strategic

E) Historical_Informational

Do not return anything except a single letter (A–E).

**Unified Fallback Prompt – Sentiment (Letter-Only)**

Return ONLY one letter corresponding to the sentiment within the given frame:

A) Negative

B) Neutral

C) Positive

Do not return anything except a single letter (A–C).

## 4. Experimental Design

The experimental design is a mixed-methods approach, combining large-scale quantitative comparison with fine-grained qualitative analysis. The overall workflow is illustrated in Diagram 3. Quantitatively, the study examines aggregate sentiment distributions and model-to-model divergence across the full corpus, enabling the identification of systematic patterns rather than isolated cases. Qualitatively, selected model rationales and frame-conditioned outputs are analyzed to contextualize these patterns and to interpret how evaluative meaning is constructed at the level of individual headlines. This integration allows the analysis to move beyond numerical comparison toward an interpretive reading of algorithmic behavior.

**Diagram 4. Overview of the Experimental Workflow**

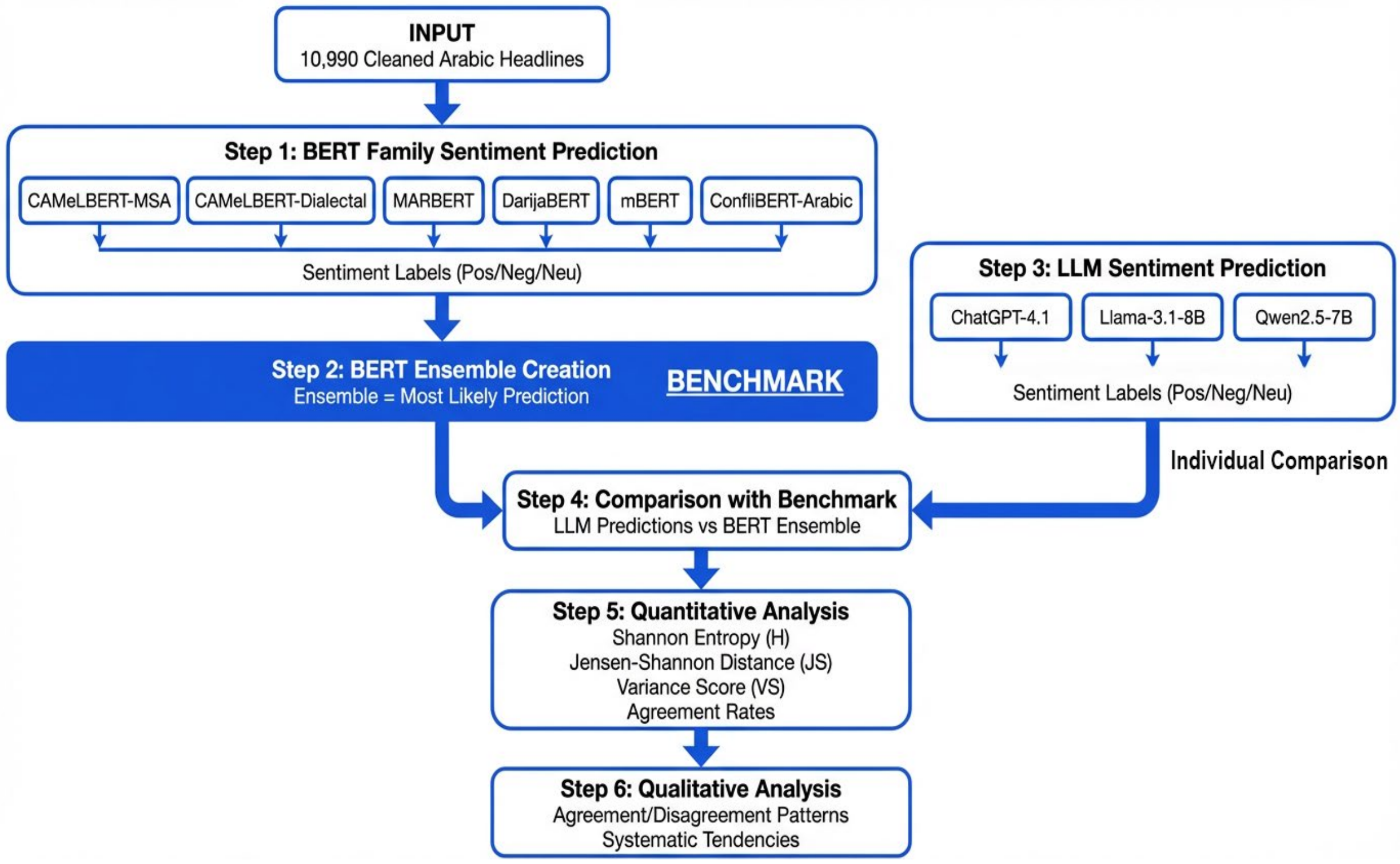

Source: Developed by the authors to outline the end-to-end experimental process.

### 4.1. Overview of the Workflow

The experiment operates on the corpus of 10,990 cleaned Arabic news headlines. Each headline is processed independently by all models under evaluation. No model has access to the predictions of other systems at inference time, ensuring the independence of outputs.

In the first stage, sentiment labels are generated by the six Arabic BERT models. These individual predictions are then aggregated to construct a BERT-based ensemble benchmark, defined as the most frequently predicted sentiment label across the six models for each headline. This ensemble serves as a stable, linguistically grounded reference point for comparative analysis.

In parallel, sentiment predictions are generated by the three LLMs using the unified two-stage prompting protocol. Each LLM assigns a sentiment label independently of the BERT ensemble, allowing for a direct comparison between generalist generative models and specialized encoder-based classifiers.

### 4.2. Operational Inference Protocol

LLM inference was conducted using a sequential batch-processing procedure applied to the cleaned headline text. To mitigate API rate limits and transient failures, a fixed inter-request delay of 0.2 seconds was introduced. Exception handling mechanisms logged failing indices, paused execution briefly, and resumed processing without terminating the overall run.

To ensure robustness and reproducibility, intermediate results were checkpointed to disk at regular intervals (every 300 headlines). This checkpointing strategy enabled resumable execution and ensured that all stored predictions correspond to a single, auditable inference pass applied consistently across the full corpus.

## 5. Evaluation Metrics

To quantitatively assess and compare sentiment distributions produced by the evaluated models, we employ a set of complementary information-theoretic and distributional metrics. These measures are designed to capture uncertainty, similarity, and deviation in model outputs without assuming the existence of a human-annotated gold standard.

### 5.1. Shannon Entropy (H)

Entropy is a standard measure of uncertainty for discrete distribution. For a discrete random variable X taking values {x1, …, xn} with probabilities P(xi), the (Shannon) entropy is defined as:

$$H(X) = -\sum_{i=1}^{n} P(x_i) log^2 P(x_i).$$

In this work, P(x_i) denotes the empirical probability of sentiment label i (Positive, Negative, or Neutral), computed from the model outputs as P(x_i) = n_i / N, where n_i is the count of headlines assigned to label i and N is the total number of headlines. Using log base 2 expresses entropy in bits (Shannon 1948). In implementation, zero-probability terms are omitted to avoid evaluating log(0).

### 5.2. Jensen–Shannon distance (JS)

To compare the similarity of two sentiment distributions P and Q, we use the Jensen–Shannon distance as implemented in SciPy (which returns the square root of the Jensen–Shannon divergence). Let M = ½(P + Q). The Jensen–Shannon divergence is:

$$JSD(P \parallel Q) = \frac{1}{2} D_{KL}(P \parallel M) + \frac{1}{2} D_{KL}(Q \parallel M), \quad M = \frac{1}{2}(P + Q)$$

The Kullback–Leibler (KL) divergence is defined as (Kullback and Leibler 1951):

$$D_K L(P \parallel Q) = \sum_{i=1}^{n} P(x_i) log\big(P(x_i)/Q(x_i)\big).$$

The Jensen–Shannon distance reported in our tables is then:

$$JS(P, Q) = \sqrt{\big(JSD(P \parallel Q)\big)}.$$

We compute JS(P, Q) using scipy.spatial.distance.jensenshannon with its default logarithm base (natural log unless otherwise specified), ensuring direct alignment with the code used to produce the results.

### 5.3. Variance Score (VS) (distributional deviation in percentage space)

To quantify how atypical a model's overall sentiment distribution is relative to the set of models, we define a variance-like deviation score in percentage space over the three sentiment classes. For each model m, let

$$r_m = \big(r_{m,pos}, r_{m,neg}, r_{m,neu}\big)$$

denote the model’s sentiment rates expressed as percentages (summing to 100). Let K be the number of models and define the mean rates across models as

$$r^{-} = (1/K) \sum_{m=1}^{K} r_m.$$

We then compute the Variance Score for model m as:

$$VS(m) = \sum_{c \in pos,neg,neu} \left(r_{m,c} - \bar{r}_c\right)^2.$$

Higher VS values indicate stronger deviation from the average model behavior. This definition matches the implementation used to compute the “Variance Score” column in the results table.

### 5.4. BERT-family benchmark selection

For cross-model comparison, we adopt the Arabic BERT-family models (and their ensemble) as the primary benchmark because they provide a specialized, linguistically grounded reference point for Arabic sentiment classification. Unlike autoregressive LLMs that are trained to predict the next token, BERT is an encoder-only, deeply bidirectional Transformer trained with masked-language modeling, enabling it to form contextual representations by conditioning on both left and right context simultaneously—an inductive bias that is particularly well-suited for sentence- and headline-level classification (Devlin et al. 2019).

This choice is especially motivated by the properties of Arabic: its morphological richness, orthographic ambiguity, and dialectal variation create substantial challenges for robust text modeling (Habash 2010). Arabic-specific BERT variants (e.g., AraBERT, CAMeLBERT, MARBERT) were introduced precisely to address these challenges by pre-training on large Arabic corpora and, in several cases, explicitly incorporating language-variant coverage (MSA vs. dialectal Arabic) and domain-relevant data distributions (Antoun, Baly, and Hajj 2020; Inoue et al. 2021; Abdul-Mageed, Elmadany, and Nagoudi 2021).

Accordingly, using the BERT-family ensemble as a benchmark anchors distributional comparisons (e.g., Jensen–Shannon distances) to a consensus baseline derived from models optimized for Arabic language understanding and discriminative classification, rather than to a single model's idiosyncratic behavior.

## Results

This section presents the core findings of the comparative sentiment analysis, which sought to understand how different Artificial Intelligence models "read" and "classify" the underlying sentiment in 10,990 Arabic news headlines concerning the Gaza conflict. The results demonstrate that the models do not converge on a single view; rather, each model possesses a distinct "interpretive lens" that leads to systematic variations in classifying headlines as "Negative," "Neutral," or "Positive."

1. **Aggregate Sentiment Distribution: The Models' Different Lenses (RQ1a)**

The analysis of the aggregate sentiment distribution reveals a clear split between model families, reflecting a fundamental difference in how each model approaches textual understanding.

Table 4: Aggregate Sentiment Distribution Across Models

| **Index** | **Model** | **Negative_%** | **Neutral_%** | **Positive_%** | **Entropy** |
|---|---|---|---|---|---|
| **0** | CAMeLBERT-MSA | 47.29 | 50.25 | 2.46 | 1.1415 |
| **1** | CAMeLBERT-DA | 55.41 | 41.90 | 2.69 | 1.1382 |
| **2** | MARBERT | 4.05 | 95.80 | 0.15 | 0.2605 |
| **3** | DarijaBERT | 36.30 | 55.17 | 8.53 | 1.3069 |
| **4** | mBERT | 27.97 | 54.54 | 17.49 | 1.4311 |
| **5** | BERT-Ensemble | 32.57 | 66.35 | 1.08 | 0.9905 |
| **6** | GPT-4.1 | 65.52 | 22.58 | 11.90 | 1.2499 |
| **7** | Qwen2.5-7B | 69.52 | 29.11 | 1.37 | 0.9679 |
| **8** | LLaMA-3.1-8B | 98.94 | 1.04 | 0.02 | 0.0859 |

**BERT Models: The Preference for Neutrality (The Literal Reader)**

The Arabic BERT-family models exhibit a strong tendency toward classifying headlines as Neutral. This inclination is most pronounced in MARBERT, which classified 95.80% of all headlines as neutral.

To understand this trend, we use the Entropy (H) measure, a statistical metric that reflects the diversity or balance of classifications.

- Low Entropy (e.g., H = 0.2605 for MARBERT) signifies a "monochromatic view," meaning the model tends to classify almost everything into a single category (Neutrality, in this case).
- High Entropy (e.g., H = 1.4311 for mBERT) signifies a "diverse view," meaning the model distributes its classifications more evenly across Negative, Neutral, and Positive labels.

The BERT models' preference for neutrality can be interpreted as them acting as a "Literal Reader" that focuses strictly on the words and sentences without applying a broad evaluative judgment. They perceive most news headlines as mere factual reporting, hence classifying them as neutral.

**Large Language Models (LLMs): The Preference for Negativity (The Evaluative Reader)**

In stark contrast, Large Language Models (LLMs) display radically different behavior. Both GPT-4.1 and Qwen2.5-7B tend to classify the majority of headlines as Negative (65.52% and 69.52%, respectively).

LLaMA-3.1-8B represents the most extreme case, classifying 98.94% of headlines as negative, resulting in the lowest entropy score (H = 0.0859). This suggests the model views everything related to the conflict as overwhelmingly negative.

This trend can be explained by the LLMs acting as an "Evaluative Reader"; they do not just read the words but invoke the broader context (stored in their massive training data) that links the conflict to harm and loss. Consequently, they systematically amplify negative sentiment.

Table 5: Jensen–Shannon Distance (JSD) Relative to the Benchmark

| Index | Model | JS_vs_Ensemble |
|---|---|---|
| 0 | CAMeLBERT-MSA | 0.1174 |
| 1 | CAMeLBERT-DA | 0.1752 |
| 2 | MARBERT | 0.2805 |
| 3 | DarijaBERT | 0.1395 |
| 4 | mBERT | 0.2180 |
| 5 | GPT-4.1 | 0.3329 |
| 6 | Qwen2.5-7B | 0.2675 |
| 7 | LLaMA-3.1-8B | 0.5457 |

**Jensen–Shannon Distance (JSD): Measuring Distance from Consensus**

The Jensen–Shannon Distance (JSD) is a metric that quantifies the difference between a model's sentiment distribution and a reference distribution (in this case, the "consensus" of the BERT models).

- A low JSD value (e.g., 0.1174 for CAMeLBERT-MSA) means the model's view is very close to the consensus.
- A high JSD value means the model's view is far from the consensus.

The results show that LLMs diverge significantly from the consensus, with LLaMA-3.1-8B recording the highest distance (JSD = 0.5457), indicating its sentiment view is the furthest from the reference.

Table 6: Variance Score Analysis – Identifying Outliers

| **Index** | **Model** | **Variance_Score** |
|---|---|---|
| **0** | CAMeLBERT-MSA | 0.002414 |
| **1** | CAMeLBERT-DA | 0.007117 |
| **2** | MARBERT | 0.446034 |
| **3** | DarijaBERT | 0.024236 |
| **4** | mBERT | 0.064820 |
| **5** | GPT-4.1 | 0.089481 |
| **6** | Qwen2.5-7B | 0.074609 |
| **7** | LLaMA-3.1-8B | 0.460702 |

The Variance Score is a measure that identifies models that represent outliers or extreme cases in sentiment classification. This analysis highlights two models as the strongest outliers:

- MARBERT (Variance Score 0.4460): The model that is extreme in Neutrality.
- LLaMA-3.1-8B (Variance Score 0.4607): The model that is extreme in Negativity.

These metrics confirm that the disagreement among models is not random but systematic and directional: some models are predisposed to favor neutrality, while others are predisposed to amplify negativity.

**Frame-Conditioned Sentiment Patterns: Does the Model Care About Context? (RQ1b)**

To understand why the models differ, we analyzed how sentiment classification changes when the narrative frame of the news (i.e., the topic focus, such as "Humanitarian" or "Security") changes. This analysis reveals the extent of the model's contextual awareness.

Table 7: Frame-Conditioned Sentiment Distribution for Large Language Models

| Model | Frame | Negative_% | Neutral_% | Positive_% | Entropy | N |
|---|---|---|---|---|---|---|
| GPT-4.1 | Historical_Informational | 6.42 | 91.97 | 1.61 | 0.4612 | 685 |
| GPT-4.1 | Humanitarian | 80.80 | 1.35 | 17.85 | 0.7764 | 1255 |
| GPT-4.1 | Legal_Accountability | 84.64 | 8.21 | 7.15 | 0.7720 | 755 |
| GPT-4.1 | Political_Strategic | 54.08 | 37.24 | 8.68 | 1.3163 | 4321 |
| GPT-4.1 | Security | 79.64 | 4.17 | 16.19 | 0.8778 | 3984 |
| Qwen2.5-7B | Historical_Informational | 9.23 | 90.38 | 0.38 | 0.4800 | 260 |
| Qwen2.5-7B | Humanitarian | 76.17 | 18.46 | 5.38 | 0.9758 | 1544 |
| Qwen2.5-7B | Legal_Accountability | 72.67 | 26.45 | 0.87 | 0.9018 | 344 |
| Qwen2.5-7B | Political_Strategic | 37.84 | 60.50 | 1.65 | 1.0670 | 1995 |
| Qwen2.5-7B | Security | 79.36 | 20.18 | 0.45 | 0.7658 | 6857 |
| LLaMA-3.1-8B | Historical_Informational | 97.77 | 2.20 | 0.03 | 0.1562 | 3813 |
| LLaMA-3.1-8B | Humanitarian | 99.87 | 0.13 | 0.00 | 0.0146 | 1503 |
| LLaMA-3.1-8B | Political_Strategic | 62.50 | 37.50 | 0.00 | 0.9544 | 8 |
| LLaMA-3.1-8B | Security | 100.00 | 0.00 | 0.00 | 0.0000 | 7 |

**GPT-4.1: A Context-Sensitive Model**

GPT-4.1 demonstrates high sensitivity to the narrative frame. When the news is Historical/Informational (reporting facts without judgment), the model classifies it as overwhelmingly Neutral (91.97%). However, when the frame shifts to Legal Accountability or Security, the negative classification jumps to over 80%.

Most tellingly, the Humanitarian frame: although predominantly negative (80.80%), it records the highest proportion of Positive sentiment (17.85%). This likely reflects the model's ability to discern positive signals related to aid and relief efforts within the crisis context.

This indicates that GPT-4.1 possesses contextual awareness that allows it to adjust its sentiment judgment based on the type of story.

### LLaMA-3.1-8B: A Context-Insensitive Model

In contrast, LLaMA-3.1-8B shows an extreme collapse in classification across all frames. Nearly all frames, even descriptive ones like "Historical/Informational," are classified as overwhelmingly negative (97.77% or more). This means the narrative frame has virtually no influence on the model's judgment, suggesting that this model applies a fixed evaluative judgment to the subject matter as a whole, regardless of the story's details.

### Summary of Findings

These results provide strong evidence that sentiment classification in conflict-related news is critically dependent on the model employed.

- BERT Models (the Literal Readers) tend to preserve descriptive neutrality and are sensitive to direct lexical cues.
- Large Language Models (the Evaluative Readers), particularly LLaMA-3.1, systematically amplify negative sentiment, suggesting they encode broader moral and contextual judgments in their classification.

The frame analysis confirms that these differences are rooted in divergent interpretations of the narrative structure of the news, rather than just surface-level lexical variation. This raises important questions about the reliability of these models for sensitive media content analysis.

## Discussion

The systematic divergence observed between the Large Language Models (LLMs) and the fine-tuned BERT models is a critical finding that challenges conventional wisdom in NLP application. The fact that LLMs, particularly the highly aggressive LLaMA-3.1-8B (which assigned up to 98.94% negative sentiment), are the most extreme in their classification, while the specialized BERT models are the most conservative (neutral), suggests a fundamental difference in how they encode and process contextual nuance in high-stakes geopolitical discourse.

The LLMs' tendency toward high negative classification is likely a result of their vast, generalist training corpora, which have imbued them with a sophisticated understanding of the real-world implications of terms like "security operation," "legal accountability," and "casualty." When prompted, the LLMs can leverage this world knowledge to infer the inherent negative valence of a conflict-related headline, even if the language is factually descriptive. This supports the conclusion that the LLM class captures the "inherently negative and polarized emotional structure" of the coverage.

The distinct interpretive lenses of the model architectures can be summarized as follows:

Table 8: What Each Model Sees: An Interpretive Comparison

| Model Type | What it Prioritizes | What it Misses / Bias |
|---|---|---|
| BERT (Fine-Tuned) | Lexical polarity (explicit sentiment words) and local linguistic patterns. | Implied violence and broad geopolitical context; defaults to neutrality. |
| MARBERT | Social media tone and colloquialisms; extreme focus on lexical cues. | Institutional framing and formal news logic; extreme bias toward Neutrality (95.80%). |
| GPT-4.1 | World knowledge and contextual implication; high sensitivity to narrative frames (e.g., Humanitarian vs. Security). | Prompt sensitivity; reproducibility can vary across prompts/model updates. |
| Qwen2.5-7B | Broad multilingual competence; strong instruction-following; intermediate position between GPT-4.1 and BERT. | Arabic nuance may depend heavily on the specific checkpoint; more conservative than GPT-4.1 in assigning non-neutral labels. |
| LLaMA-3.1-8B | General-purpose contextual inference; strong pattern completion; extreme focus on the negative valence of the conflict topic. | Contextual nuance; extreme bias toward Negativity (98.94%); frame distinctions exert minimal influence. |

This table visually reinforces the study's central argument that the choice of model is a choice of interpretive framework, each with its own inherent biases and blind spots.

Conversely, the extreme neutrality of models like MARBERT (95.80% Neutral) suggests a failure of the fine-tuning process to adequately sensitize the model to the specific, often implicit, sentiment cues of conflict reporting. While these models are excellent at general Arabic sentiment tasks, their training data may not have contained enough examples where conflict-related terms were explicitly labeled as negative, leading them to default to the safe, neutral classification when faced with ambiguous or descriptive headlines. This indicates a significant limitation in the application of standard fine-tuned BERT models to complex geopolitical analysis.

**Why Neutral Is Not Neutral in Conflict Reporting**

The high rate of algorithmic neutrality observed in the BERT models demands a critical theoretical interpretation. In the context of conflict, algorithmic neutrality is not equivalent to journalistic objectivity or true media neutrality. Instead, this computational default can be interpreted as a form of algorithmic erasure or depoliticization. By classifying headlines describing violence, casualties, or political tension as merely "neutral," the model effectively normalizes the underlying conflict and its consequences. This computational tendency can inadvertently mask the emotional and political valence inherent in the discourse, thus failing to capture the media logic of crisis reporting. The LLMs' ability to break this pattern and assign a negative label to such content suggests a more robust capacity to infer the critical, non-neutral context of geopolitical events.

The qualitative analysis of the LLMs' rationales is a significant contribution to the field. By linking sentiment polarity to specific narrative frames (Table 4 in the Results section), we move beyond a simple classification score to an interpretable understanding of the algorithmic decision-making process. This capability transforms the LLM from a black-box classifier into a valuable analytical tool for deconstructing media frames, allowing researchers to link quantitative sentiment scores to qualitative narrative content.

**Limitations and Epistemological Design Choices**

While this study provides a robust comparative analysis, it is important to acknowledge the deliberate design choices that shape its scope and interpretation. The primary limitations are:

Absence of a Human-Coded Gold Standard: The study intentionally foregoes a human-annotated "gold standard" for sentiment. This choice is rooted in the epistemological belief that sentiment in a highly contested geopolitical context is inherently subjective and contested, making a single, definitive human label misleading.

Reliance on a Single LLM for Framing: The qualitative analysis of narrative frames is based solely on the rationales generated by GPT-4.1. While this provides a consistent interpretive lens, it does not account for potential variance in framing interpretation across all LLMs (e.g., LLaMA-3.1-8B's near-total collapse to negativity suggests its frame analysis would be less informative).

Focus on Headlines Only: The analysis is restricted to news headlines, which are highly condensed linguistic units. This excludes the richer, more detailed contextual information available in the full body text of the articles.

Crucially, these are not methodological flaws, but deliberate design choices aligned with the study's epistemological stance. The goal is not to measure accuracy against a fixed truth, but to analyze the variance and interpretive acts of the models themselves. The limitations thus serve to sharpen the focus on the core research question of algorithmic discrepancy.

**Conclusion**

This study examined how multiple Large Language Models (GPT-4.1, Qwen2.5-7B, and LLaMA-3.1-8B) and fine-tuned Arabic BERT models classify sentiment in more than 10,900 Arabic news headlines about the Gaza conflict, focusing on their algorithmic discrepancy rather than traditional accuracy against a single gold standard. The findings demonstrate that disagreements between models are both substantial and systematic, reflecting deeper differences in how each architecture processes contextual nuance, encodes latent biases, and maps conflict-related language onto discrete sentiment categories.

The systematic variance—specifically the LLMs' tendency toward negativity and the BERT models' tendency toward neutrality—highlights the profound impact of model architecture and training data on the interpretation of geopolitical discourse. We conclude that neither model class provides a singular "ground truth." Instead, the comparison of their outputs offers a richer, more triangulated insight into media framing and the algorithmic mediation of war narratives. Future research should continue to leverage the complementary strengths of these models, using LLMs for qualitative frame analysis and BERT models for high-throughput, specialized lexical detection.